\documentclass{article}
\usepackage{ijcai19}

\usepackage{times}
\usepackage{soul}
\usepackage{url}
\usepackage[hidelinks]{hyperref}
\usepackage[utf8]{inputenc}
\usepackage[small]{caption}
\usepackage{graphicx}
\usepackage{amsmath}
\usepackage{booktabs}
\usepackage{algorithm}
\usepackage{algorithmic}

\usepackage{setspace}
\usepackage{bm}
\usepackage{amsmath}
\usepackage{latexsym}
\usepackage{enumerate}
\graphicspath{{./pic/}}
\DeclareGraphicsExtensions{.pdf}

\title{RTHN: A RNN-Transformer Hierarchical Network for Emotion Cause Extraction}

\author{
Rui Xia\and
Mengran Zhang\and
Zixiang Ding
\affiliations
School of Computer Science and Engineering, 
\affiliations
Nanjing University of Science and Technology, China
\emails
\{rxia, zhangmengran, dingzixiang\}@njust.edu.cn
}
\begin{document}

\maketitle

\begin{abstract}
The emotion cause extraction (ECE) task aims at discovering the potential causes behind a certain emotion expression in a document. Techniques including rule-based methods, traditional machine learning methods and deep neural networks have been proposed to solve this task. However, most of the previous work considered ECE as a set of independent clause classification problems and ignored the relations between multiple clauses in a document. In this work, we propose a joint emotion cause extraction framework, named RNN-Transformer Hierarchical Network (RTHN), to encode and classify multiple clauses synchronously. RTHN is composed of a lower word-level encoder based on RNNs to encode multiple words in each clause, and an upper clause-level encoder based on Transformer to learn the correlation between multiple clauses in a document. We furthermore propose ways to encode the relative position and global predication information into Transformer that can capture the causality between clauses and make RTHN more efficient. We finally
achieve the best performance among 12 compared systems and improve the F1 score of the state-of-the-art from 72.69\% to 76.77\%.
\end{abstract}

\section{Introduction}
Emotion cause extraction (ECE) is a fine-grained task of emotion analysis, which aims at discovering the potential causes behind a certain emotion expression in the text. The ECE task was first proposed and defined as a word-level sequence labeling problem in \cite{lee2010text}. To solve the shortcomings of describing emotion cause at word/phrase level, \cite{gui2016event} released a new corpus and re-formalized the ECE task as a clause-level classification problem. This corpus has received much attention in the following study and has become a benchmark dataset for ECE research. Figure 1 gives an example to the annotation of their corpus. In this example, a document is composed of five clauses. The emotion expression “happy” is contained in Clause $c_4$ and the corresponding cause “the thief was caught” is contained in Clause $c_3$ (We call Clause $c_4$ and Clause $c_3$ \emph{emotion expression clause} and \emph{emotion cause clause}, respectively). The goal of ECE is to predict for each clause in a document, whether this clause contains an emotion cause, given the annotation of the emotion expression.

Rule-based methods and traditional machine learning methods have been proposed to address this problem. In recent years, deep neural networks have also been applied to this task and achieved state-of-the-art performance \cite{cheng2017emotion,gui2017question,li2018co,ding2019independent,yu2019multiple}.
\begin{figure}[!t]
\centering
\includegraphics[width=3.3in]{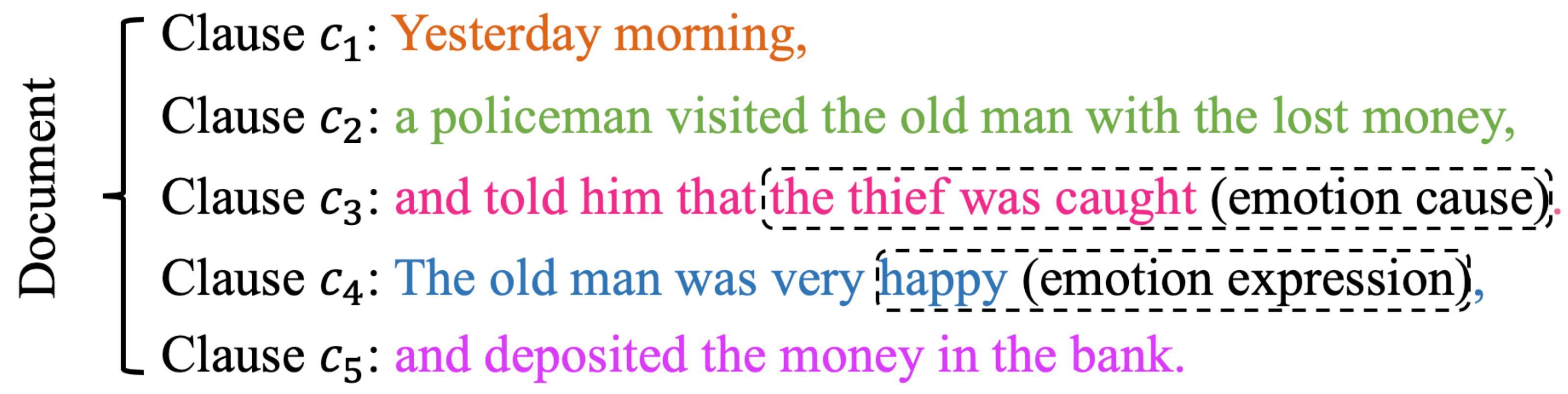}
\caption{An example of the emotion cause extraction task.}
\label{fig_sim1}
\end{figure}

However, most of the previous work considered ECE as a set of independent clause classification problems. For example, under this framework, Example 1 will be formalized as five independent classification tasks (Clause $c_1$ to Clause $c_5$). Although this framework was straight-forward, it ignores the relations between multiple clauses for emotion cause inference. There are two types of relationships between clauses: 1) Correlation: two clauses with similar semantics are supposed to have similar probabilities being the emotion cause. 2) Causality: incorporating the information of the other clauses in the document can help infer the current clause in a global view. It was observed from the corpus \cite{gui2016event} that more than 99\% of the documents have only one or two causes. If one clause has a high probability being an emotion cause, the probabilities that other clauses being emotion causes should be reduced; conversely, if no high-confidence emotion cause clauses have been observed, the probability of predicting the current clause being an emotion cause should be increased.

In this work, we propose a hierarchical network architecture based on RNN and Transformer, named RNN-Transformer Hierarchical Network (RTHN), to model the relations between multiple clauses in a document and classify them synchronously in a joint framework. RTHN is composed of two layers: 1) The lower layer is a word-level encoder consisting of multiple RNNs, each of which corresponds to one clause, in turn encodes the words in the clause and combines them to obtain the clause representation; 2) The upper layer is a clause-level encoder based on a stacked Transformer, where the clause representations are repeatedly learned and updated by incorporating the relations between multiple clauses, and finally feed to a softmax layer for synchronous classification.

We further propose ways to encode the relative position and global prediction information which have been proven to be important features in ECE, and gain further improvements. On one hand, the attention mechanism in Transformer learns the correlation between clauses. On the other hand, the encoding of global prediction the causality between clauses.

The main contributions of this work can be summarized as follows:
\begin{enumerate}
\item We propose a new hierarchical network architecture based on RNNs and Transformer for the ECE task. To the best of our knowledge, it is the first time that Transformer has been used to solve ECE problems. It demonstrates excellent performance in learning the correlation between multiple clauses in ECE.
\item We further encode the relative position and global predication information into the Transformer framework. It can capture the causality between clauses and achieve extra improvements.
\item The effectiveness of our model is demonstrated on the benchmark ECE corpus. We finally achieve the best performance among 12 compared systems and improve the F1 score of the state-of-the-art from 72.69\% to 76.77\%\footnote{The source code can be obtained at https://github.com/NUSTM/RTHN}.
\end{enumerate}

\section{Related Work}
\cite{lee2010text} manually constructed a small-scale emotion cause corpus based on the Academia Sinica Balanced Chinese Corpus and first proposed the emotion cause extraction (ECE) task. In this corpus, the spans of both emotion expression and emotion cause were annotated and the ECE task was defined as a word-level sequence labeling problem. Subsequent research proposed either rule-based methods or machine learning methods to solve this problem based on manually designed rules or features \cite{chen2010emotion,lee2013detecting}. 

\cite{li2014text} constructed an emotion cause corpus based on Chinese microblog posts, and proposed a rule-based method to infer and extract the emotion cause by importing knowledge and theories from other fields such as sociology. \cite{gao2015emotion} and \cite{gao2015rule} further designed a set of complex rules considering a cognitive emotion model and emotions categories to extract emotion cause on this corpus. \cite{gui2014emotion} also constructed a microblog emotion cause corpus based on NLPCC 2013 emotion analysis share task, and proposed a machine learning method based on SVMs and conditional random fields (CRFs) to extract emotion causes. Similar as \cite{lee2010text}, the ECE task in these corpora was defined and evaluated as a sequence labeling problem.

There were also some individual studies that conducted ECE research on their own corpus \cite{russo2011emocause,neviarouskaya2013extracting,ghazi2015detecting,song2015detecting,yada2017bootstrap}. They normally regarded ECE as a sequence labeling problem and employed rule-based or traditional machine learning algorithms to solve it.

\cite{gui2016event} and \cite{gui2016emotion} released a Chinese emotion cause corpus from a public SINA city news and proposed a multi-kernel based method for emotion cause extraction. Different from previous corpora, the ECE task in this corpus was defined as a clause classification problem, where the goal is to predict for each clause in a document, whether this clause is an emotion cause, given the annotation of emotion expression. It was also evaluated by the clause-level Precision, Recall and F1 score metrics. This corpus has received much attention in the following study and has become a benchmark dataset for ECE research. Several traditional machine learning methods including structure representation and multi-kernel learning has been proposed in \cite{gui2016emotion,xu2017ensemble}. \cite{gui2016event} proposed a tree structure-based representation method to describe the events in emotion cause extraction on this corpus. In recent two years, deep learning techniques have also been applied to emotion cause extraction. For example, \cite{cheng2017emotion} used long short-term memory (LSTM), \cite{gui2017question} proposed a deep memory network, and \cite{li2018co} proposed a co-attention neural network, for emotion cause prediction.

Most of the above work considered the ECE task on this corpus as a set of independent clause classification tasks and ignored the relations between multiple clauses in a document. To address this, \cite{ding2019independent} converted the task to a reordered clause classification problem. The predictions of previous clauses were used as features for predicting subsequent clauses. However, their approach depends on the clause order and can only use the predictions of the previous clauses but not the subsequent clauses. In contrast, RTHN proposed in this work is a joint emotion cause extraction framework that models and classifies multiple clauses in a document synchronously.

\cite{yu2019multiple} proposed a three-level (word-phrase-clause) hierarchical network based on CNNs and LSTMs. The multiple clauses are modeled with LSTMs in their approach. By contrast, in this work Transformer is used as the clause-level encoder to model the relations between multiple clauses. We will empirically prove that Transformer has shown significantly better performance as a clause-level encoder, in comparison with RNNs.

\section{Approach}
\subsection{Overall Architecture}
In this paper, we represent a document containing multiple clauses as $d=\{c_1,...,c_i,...,c_{|d|}\}$, where $c_i$ is the $i$-th clause in $d$. Each clause $c_i$ consists of multiple words $c_i=\{w_{i,1},...,w_{i,t},...,w_{i,|c_i|}\}$.

In this work, we propose a RNN-Transformer Hierarchical Network (RTHN) to model such a “word-clause-document” hierarchical structure\footnote{The document is usually very short in the benchmark corpus \cite{gui2016event}. Most of them contain less than 20 clauses. We therefore ignore the sentence level between clause and document.}. The overall architecture of RTHN is shown in Figure 2. It contains two layers: the lower layer is a word-level encoder consisting of multiple Bi-LSTM modules, each of which corresponds to a clause (Section 3.2); the upper layer is a clause-level encoder consisting of a stacked Transformer module, where the clause representations obtained at lower layer are repeatedly updated by incorporating the relations between clauses (Section 3.3), relative position and global prediction (Section 3.4), and finally feed to a softmax layer for synchronous classification.

\subsection{Word Level Encoder Based on RNNs}
The lower layer is a word-level encoder consisting of multiple Bi-LSTMs. Each clause corresponds to a Bi-LSTM module, which accumulate the context information for each word of the clause. The hidden state of the $t$-th word in the $i$-th clause $h_{i,t}$ is obtained based on a bi-directional LSTM. A word-level attention mechanism is then adopted to get the clause representation $r_i$ by a weighted sum of hidden states of all the words in the clause. Here we omit the details of Bi-LSTM and attention for limited space.

As has mentioned in the Introduction, most of the previous studies considered ECE as a set of independent clause classification problem. This framework normally has only one encoding layer (word-level encoder). In this work, the proposed RTHN model is a 2-layer hierarchical network containing not only word-level encoders at the lower layer but also a clause-level encoder at the upper layer.
\begin{figure}[!t]
\centering
\includegraphics[width=3.3in]{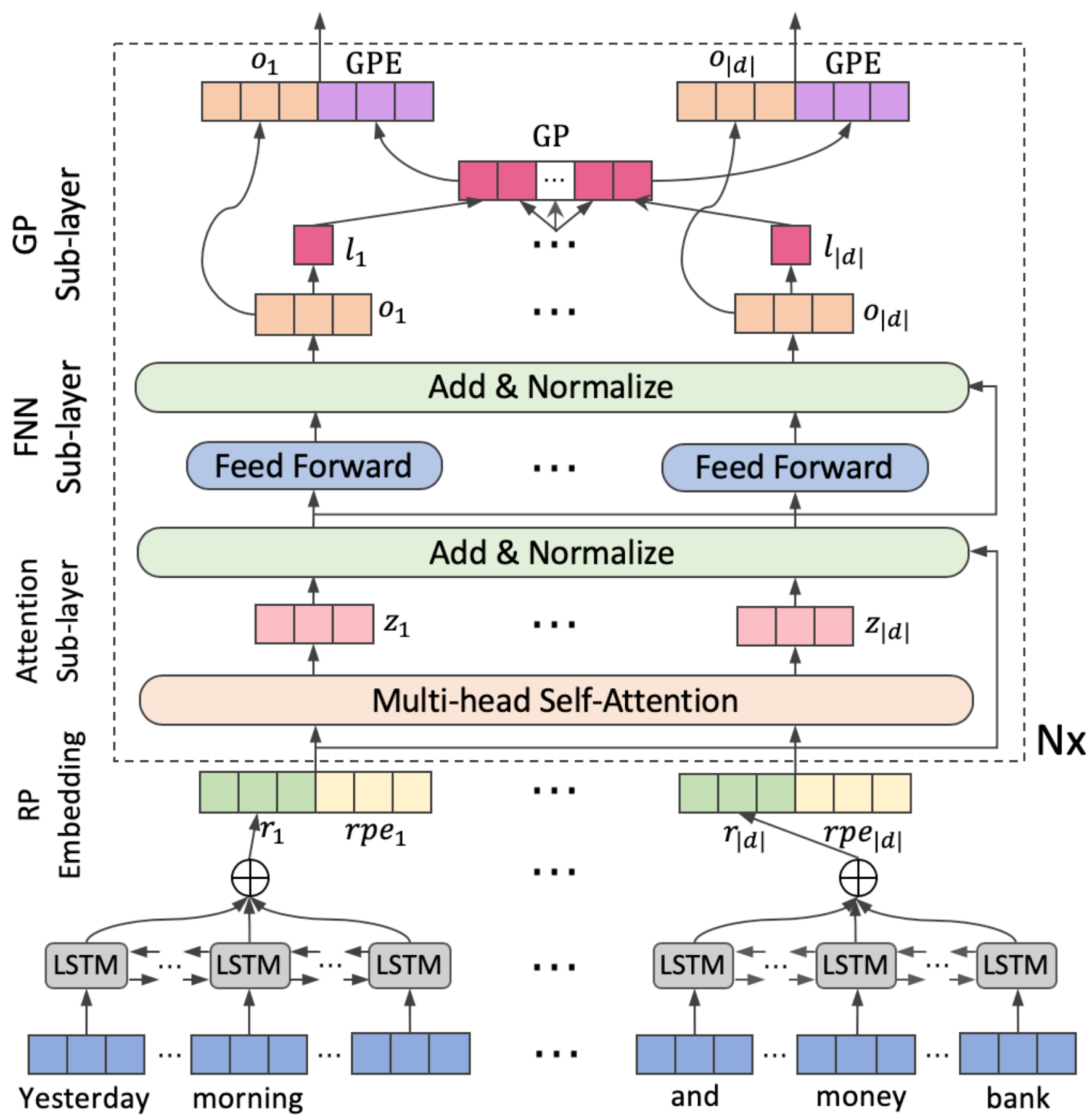}
\caption{The framework of our RTHN model.}
\label{fig_sim1}
\end{figure}

\subsection{Clause Level Encoder Based on Transformer}
In this work, Transformer \cite{vaswani2017attention} is used as the clause-level encoder at the upper layer to encode the relations between multiple clauses in a document.

The standard Transformer consists of a stack of $N$ layers. Each layer has two sub-layers: the first is a multi-head self-attention mechanism; the second is a fully connected feed-forward network.

\begin{spacing}{1.3}
\noindent\textbf{(1) Multi-head Self-attention}
\end{spacing}
\noindent The unit of multi-head self-attention mechanism is the major component in the Transformer. For each clause $c_i$, the representation $r_i$ obtained at the word-level encoder is used as the input $x_i$ by adding a positional embedding $p_i$:
\begin{equation}
\label{eqn_example}
x_i=r_i+p_i.
\end{equation}

In our setting, the Transformer contains a query vector $q_i$, a key vector $k_i$ and a value vector $v_i$ for each clause $c_i$ in the document:
\begin{gather}
\label{eqn_example}
q_i={\rm ReLU}(x_iW_Q), \\
k_i={\rm ReLU}(x_iW_K), \\
v_i={\rm ReLU}(r_iW_V),
\end{gather}
\noindent where $W_Q$, $W_K$ and $W_v$ are learnable weight matrix of query, keys and values, respectively.

For each clause $c_i$, self attention learns a set of weights $\beta_i=\{\beta_{i,1},\beta_{i,2},...,\beta_{i,|d|}\}$, which measures the extent of all the input clauses $[c_1,..,c_k,...,c_{|d|}]$ answer the query $q_i$:
\begin{equation}
\label{eqn_example}
\beta_{i,j}=\frac{{\rm exp}({q_i}\cdot k_j)}{\sum_{j^{'}}{\rm exp}({q_i\cdot k_{j^{'}})}}.
\end{equation}
The output is a weighted sum of the values of all clauses:
\begin{equation}
\label{eqn_example}
z_i=\sum_{j}\beta_{i,j}v_j.
\end{equation}
This allows that the representation of each clause can encode a global level information on all the clauses in the document, rather than rely solely on the hidden state of one clause.

Moreover, the multi-head attention is employed with the number of heads as 5.

\begin{spacing}{1.3}
\noindent\textbf{(2) Feed-Forward Network}
\end{spacing}
\noindent The attention sublayer is then followed by a fully connected Feed-Forward Network (FFN) sublayer:
\begin{equation}
\label{eqn_example}
e_i={\rm ReLU}(z_iW_1+b_1)W_2+b_2.
\end{equation}

Note that both of the above two sublayers use the residual connection followed by normalization layer at its output:
\begin{equation}
\label{eqn_example}
o_i={\rm Normalize}(e_i+x_i).
\end{equation}

As has mentioned, Transformer is a stack of $N$ layers each of which includes attention and FFN sublayers. Let $l$ denote the index of Transformer layers. The output of the previous layer will be used as the input of the next layer:
\begin{equation}
\label{eqn_example}
x_i^{(l+1)}=o_i^{(l)}.
\end{equation}

\subsection{Encoding Relative Position and Global Prediction}
In this section, we propose to further encode the relative position and global prediction which have been proven to be two explicit clues in ECE. 

\begin{spacing}{1.3}
\noindent\textbf{(1) Relative Position Embedding}
\end{spacing}
\noindent In standard Transformer, a positional encoding was already applied to represent the of position information. But what it reflects is the absolute positional information of the word in the sentence.

In our task, relative position (RP) is defined as the relative distance between the current clause and the emotion expression clause. For example, -1 is the RP of the clause left to the emotion expression clause, +2 the relative position of the second clause right to the emotion expression clause, and so on. 

Relative position is more important than absolute position in ECE, because people are more inclined to explain the causes near the emotion expression. Therefore, the clauses with smaller relative position (rather than absolute position) are more likely to be an emotion cause of the given emotion expression.

In this work, we use relative position embedding (RPE) to encode such relative position information. The RPEs for all the clauses in a document are denoted by $\{rpe_1,rpe_2,...,rpe_{|d|}\}$ where $rpe_i$ denotes the RPE of clause $c_i$. Instead of Equation (1), $rpe_i$ is concatenated to the clause representation $r_i$, as the input of the Transformer:
\begin{equation}
\label{eqn_example}
x_i=r_i\oplus rpe_i.
\end{equation}

\begin{spacing}{1.3}
\noindent\textbf{(2) Global Prediction Embedding}
\end{spacing}
\noindent As we have mentioned, there are two types of relationships between clauses: correlation and causality. In addition to using the attention mechanism in Transformer to capture the correlation between clauses, we furthermore propose to append a new global prediction sublayer to the end of each Transformer layer to introduce more causality.

Global Prediction (GP) denotes the prediction labels of all the clauses in a document. As can be observed in the ECE corpus \cite{gui2016event}, more than 99\% of the documents have only one or two causes. If one clause in the document is predicted as an emotion cause with high confidence, the probability that other clauses are predicted as emotional causes should be reduced; conversely, if there are no other emotional cause clauses with high confidence in the document, the probability that the current clause is predicted to be an emotion cause should be increased. Therefore, GP is an important clue for emotion cause extraction.

Firstly, we get the prediction label of each clause $l_i\in\{+1,-1\}$ based on the output representations $o_i$:
\begin{equation}
\label{eqn_example}
l_i\leftarrow {\rm softmax}(Wo_i+b).
\end{equation}

Secondly, we sort the predicted labels of different clauses according to their relative positions […, -2, -1, 0, +1, +2, …] and build the following global prediction vector
\begin{equation}
\label{eqn_example}
GP=[...,l_{i_{-2}},l_{i_{-1}},l_{i_{0}},l_{i_{+1}},l_{i_{+2}},...].
\end{equation}
where $l_{i_{rp}}$ denotes the prediction label of the clause at relative position $rp$ and the current position (i.e). Note that we mask the prediction at the current position to avoid potential interference. For example, if the $rp$ of current clause is +1, we let $l_{i_{+1}}=0$. $GP$ represents different combinations of all clause predictions and is then encoded by an embedding called Global Prediction Embedding $GPE$:
\begin{equation}
\label{eqn_example}
GPE={\rm Tanh}(W_{gpe}GP+b_{gpe}),
\end{equation}
where $W_{gpe}$ and $b_{gpe}$ are learnable matrix and bias.

In stacking, the average $GPE$ of previous layers is concatenated to the output representation $o_i^{(l)}$ and used as the input of the next layer’s input:
\begin{equation}
\label{eqn_example}
x_i^{(l+1)}=o_i^{(l)}\oplus Ave\_GPE^{(l)},
\end{equation}
where $Ave\_GPE^{(l)}=\frac{1}{l}\sum_{l}GPE^{(l)}$.

\subsection{Multiple Clause Classification}
After a stack of $N$ layers, we obtain the final clause representation $o_i^{(N)}$ for each clause, and employ an extra softmax function to yield the final prediction distribution
\begin{equation}
\label{eqn_example}
\hat{y}_i={\rm softmax}(W_c^{(N)}o_i^{(N)}+b_c^{(N)}).
\end{equation}

The training objective is to minimize the cross-entropy loss across all the clauses:
\begin{equation}
\label{eqn_example}
Loss=-\sum_{d\in Corpus}\sum_{i=1}^{|d|}y_i\cdot {\rm log}{(\hat{y}}_i)+\lambda||\theta||^2,
\end{equation}
where $y_i$ is the ground-truth distribution of clause $c_i$. A L2-norm regulation is also adopted with $\lambda$ denoting the tradeoff weight.

\section{Experiments}
\subsection{Dataset and Experimental Settings}
We evaluate our RTHN model on the benchmark ECE corpus \cite{gui2016event}, which was the mostly used corpus for emotion cause extraction. The same as \cite{gui2017question}, we randomly divide the data with the proportion of 9:1, with 9 folds as training data and remaining 1 fold as testing data. The following results are reported in terms of an average of 10-fold cross-validation. The performance measures are Precision (P), Recall (R), and F1 all defined at clause level.

We use the word embedding provided by NLPCC. It was pre-trained on a 1.1 million Chinese Weibo corpora with the word2vec toolkit \cite{mikolov2013distributed}. Similar performance can be obtained by using the embedding in \cite{gui2017question}. The dimension of word embedding, RP embedding and GP embedding is set to be 200, 50 and 50, respectively. The hidden units of LSTM in word-level encoder is set to be 100. The dimension of the hidden states in Tranformer is 200, and the dimensions of query, key and value are 250, 250, and 200 repectively.

The maximum numbers of words in each clause and clauses in each document are set to be 75 and 45, respectively. The network is trained based on the Adam optimizer with a mini-batch size 32 and a learning rate 0.005.
\subsection{Compared Systems}
We compare our model with the following 12 baseline systems:
\begin{enumerate}[1)]
\item RB is a rule based method \cite{lee2010text};
\item CB is common-sense based method \cite{russo2011emocause};
\item RB+CB is a combination of RB and CB;
\item RB+CB+SVM is a SVM classifier trained on features including rules \cite{lee2010text} and Chinese Emotion Cognition Lexicon \cite{xu2017ensemble}; 
\item Ngrams+SVM denotes a SVM classifier that uses the unigram, bigram and trigram features. It was a baseline system in \cite{gui2017question}; 
\item Multi-kernel is a multi-kernel based method proposed in \cite{gui2016event};
\item Word2vec+SVM denotes a SVM classifier using word embeddings learned by Word2vec as features; 
\item CNN is the basic convolutional neural network proposed by \cite{kim2014convolutional};
\item Memnet is convolutional multiple-slot deep memory network proposed by \cite{gui2017question};
\item CANN is a co-attention neural network model with emotional context awareness \cite{li2018co};
\item PAE-DGL is a reordered prediction model that incorprates relative position information and dynamic global label \cite{ding2019independent};
\item HCS is a CNN-RNN based three-level hierarchical network based clause selection \cite{yu2019multiple}.
\end{enumerate}

\subsection{Main Results}

\begin{table}
\centering
\renewcommand\tabcolsep{4.0pt} 
\begin{tabular}{c|ccc}
\hline
& P & R & F1  \\
\hline
RB \cite{lee2010text} & 0.6747 & 0.4287 & 0.5243 \\
CB \cite{russo2011emocause} & 0.2672 & 0.7130 & 0.3887 \\
RB+CB & 0.5435 & 0.5307 & 0.5370 \\
RB+CB+SVM & 0.5921 & 0.5307 & 0.5597 \\
Ngrams+SVM & 0.4200 & 0.4375 & 0.4285 \\
Word2vec+SVM & 0.4301 & 0.4233 & 0.4136 \\
Multi-Kernel \cite{gui2016event} & 0.6588 & 0.6927 & 0.6752 \\
\hline
CNN \cite{kim2014convolutional} & 0.6215 & 0.5944 & 0.6076 \\
Memnet \cite{gui2017question} & 0.7076 & 0.6838 & 0.6955 \\
CANN \cite{li2018co} & 0.7721 & 0.6891 & 0.7266 \\
PAE-DGL \cite{ding2019independent} & 0.7619	 & 0.6908 & 0.7242 \\
HCS \cite{yu2019multiple} & 0.7388 & 0.7154 & 0.7269 \\
\hline
RTHN (layer 1) & 0.7696 & 0.7333 & 0.7501 \\
RTHN (layer 2) & 0.7644 & 0.7566 & 0.7601 \\
RTHN (layer 3) & 0.7697 & 0.7662 & {\bf 0.7677} \\
RTHN (layer 4) & 0.7604 & 0.7699 & 0.7646 \\
RTHN (layer 5) & 0.7592 & 0.7684 & 0.7634 \\
\hline
\end{tabular}
\caption{ Performance of our RTHN model and the other baseline systems on the emotion cause corpus \protect\cite{gui2016event}.}
\label{tab:plain}
\end{table}

The past clause-level approaches regarded the ECE task as a set of independent clause classification problems. By observing the corpus, we found that the proportions of emotion cause clauses and non-emotion-cause clauses were 18.36\% and 81.64\%, respectively. It is a serious class-imbalance classification problem and the model tends to predict the clause as non-emotion-cause more often. This is also the reason why their Recall scores were quite low (the highest was 0.6908).

By contrast, it can found in Table 1 that the Recall scores of the hierarchical models (HCS and RTHN) are significantly higher than pervious methods.  This is because they can capture the relations of multiple clauses which help inferring the current clause. For example, if no other clauses in a document have been detected as an emotion cause, the model will increase the probability of the current clause being predicted as an emotion cause. This finally increases the Recall score. In particular, RTHN achieves a much higher Recall score (0.7699) than other methods (the improvement is more than 7\% over the clause-level methods including CANN and PAE-DGL, and more than 5\% over HCS), but without reducing the Precision score (only slightly lower than CANN but still higher than the other baselines). 

We also plot the Precision-Recall (PR) Curves of three methods (CANN, PAE-DGL and RTHN) in Figure 3. It can be seen that the PR curve of RTHN is basically at the top-right of the other curves for most cases, and the area under RTHN PR curve is also significantly larger than the others. It further proves the superiority of the overall performance of our RTHN model.

In Figure 4, We display the attention weights learned by Transformer, by using the example in Figure 1 as a test document. The height of the $j$-th column for the $i$-th clause denotes the weight of clause $c_j$ in representing the clause $c_i$: $\beta_{i,j}$ (see Equation 5). Figure 4 can be observed from two angles.

Firstly, for each clause, the weight at the emotion expression clause is the largest and gradually becomes smaller towards both sides. This shows that Transformer can automatically capture the relative distance information: the smaller the relative distance, the larger the weight assigned. The distribution of weights looks similar to a normal distribution centered on the emotion expression clause. 

Secondly, by observing different clauses, we can find that the clause with higher probability being an emotion cause tends to have more concentrated distribution; On the contrary, clauses with smaller probabilities being an emotion cause tend to have more uniform distribution. In Figure 4 the weight distribution of clause $c_3$ is the most concentrated, where clause $c_4$ is exactly the ground-truth emotion cause. The largest weight, $\beta_{3,4}$ (0.66) also happens to be the weight of the emotion cause clause and emotion expression clause. This phenomenon is very common across different documents in our experiments. It further confirms our model's effectiveness in capturing the relationships between multiple clauses in emotion cause inference.

\begin{figure}[!t]
\centering
\includegraphics[width=3.3in]{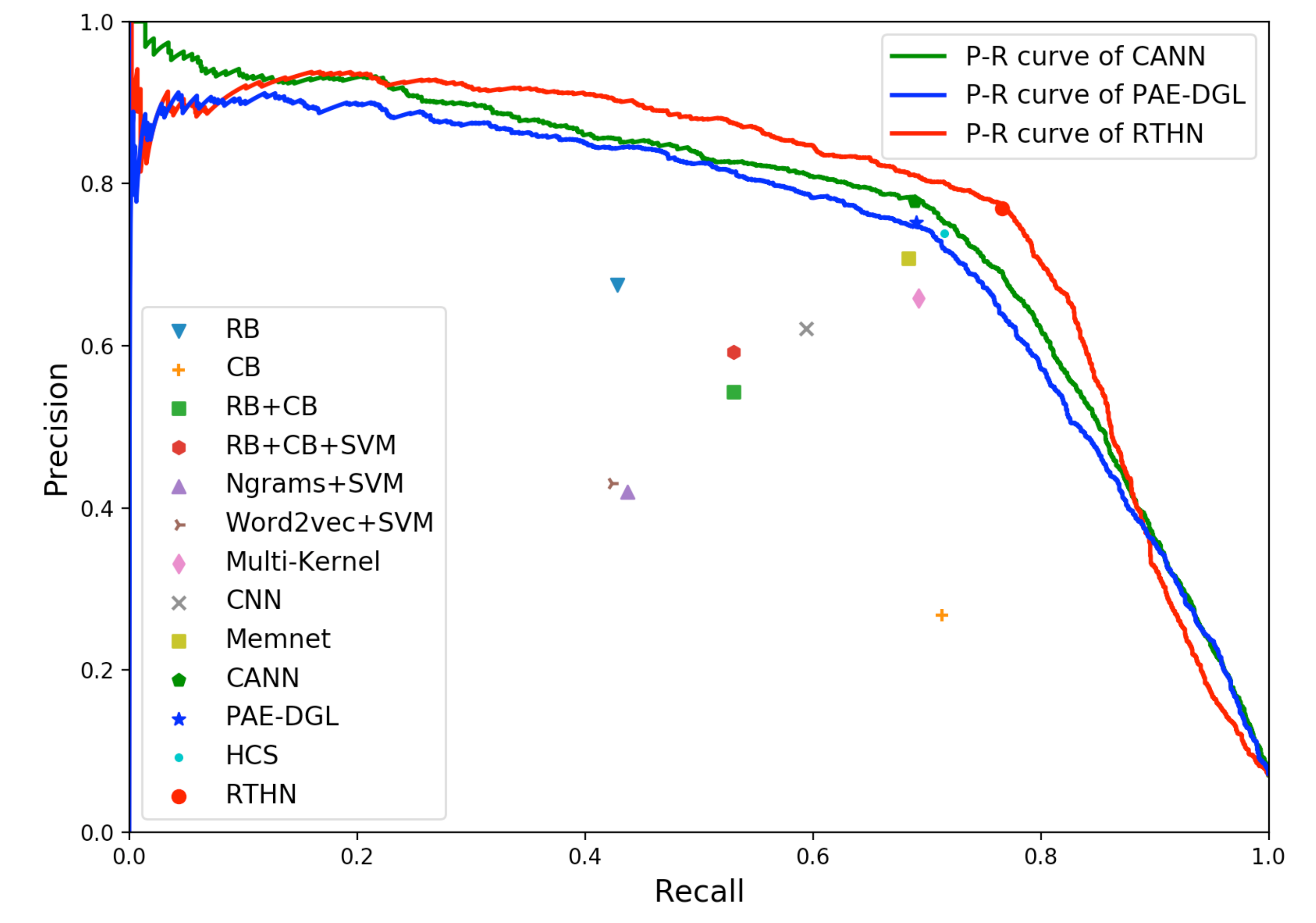}
\caption{The distribution of attention weights learned in Transformer for each clause of the example in Figure 1.}
\label{fig_sim1}
\end{figure}

\begin{figure}[!t]
\centering
\includegraphics[width=3.3in]{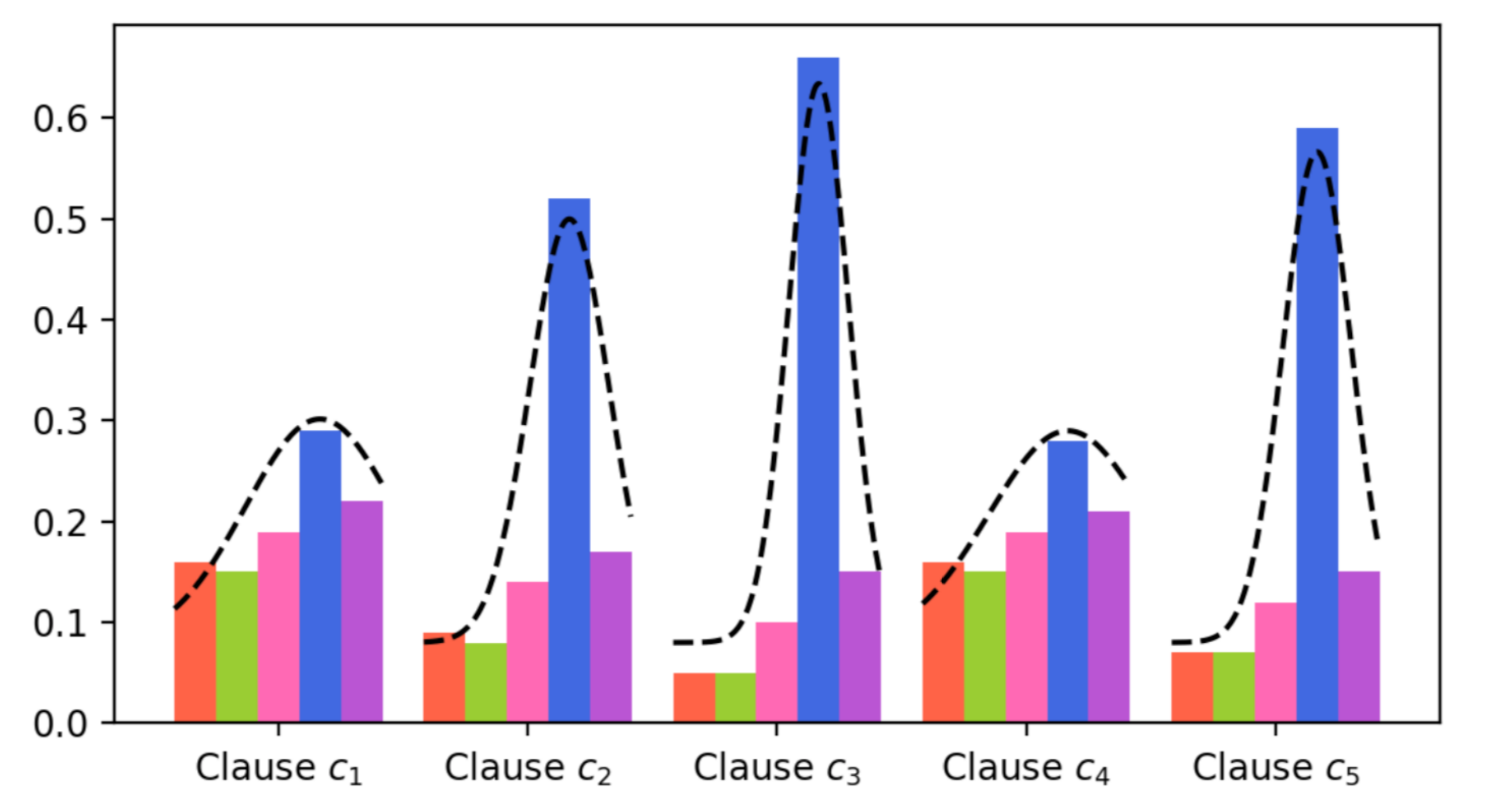}
\caption{The distribution of attention weights learned in Transformer for each clause of the example in Figure 1.}
\label{fig_sim1}
\end{figure}
\subsection{The Effectiveness of Encoding Relative Position and Global Prediction}
In order to further examine the effects of encoding relative position and global prediction in RTHN, we carry out an ablation study by designing the following RTHN variants:
\begin{itemize}
\item RTHN-No-GPE (RTHN after removing global prediction encoding);
\item RTHN-No-RPE (RTHN after removing relative position embedding);
\item RTHN-APE (RTHN using absolute position embedding instead of relative position embedding).
\end{itemize}

The results are reported in Table 2. We can observe that after reducing global prediction encoding, the F1 score of RTHN-No-GLE decreases more than 3\% (0.7314). The removal of relative position encoding results in a greater performance degradation (RTHN-No-RPE: 0.4145). By applying the absolute position embedding, RTHN-APE still perform poorly (0.5694). All these results demonstrate that RPE and GPE are important two factors in RTHN.

\subsection{Why using RNN-Transformer Combination}
In RTHN, RNNs and Transformer are used as the word-level and clause-level encoders respectively. To investigate the effectiveness of the RNN-Transformer combination, we further design two other combinations for hierarchical modeling:
\begin{itemize}
\item RRHN (RNN-RNN hierarchical network). Multiple Bi-LSTMs with attention are used as word-level encoders and a 3-layer stacked Bi-LSTMs is used as the clause-level encoder.
\item TTHN (Transformer-Transformer hierarchical network). Transformer is used as both word-level and clause-level encoders.
\end{itemize}

Note that relative position and global prediction are encoded in RRHN and TTHN the same way as that in RTHN. In Table 3, we report their performance as well as the training time on a GTX-1080Ti GPU server. It can be observed that both RRHN and TTHN perform less effectively than RTHN. But what surprised us a bit is that TTHN's performance is significantly behind RTHN and RRHN. One possible reason is that we only use a layer of Transformer in the word-level encoder. Moreover, due to the advantages of parallel computing, Transformer's training time is shorter than RNN that can only perform serial operations.
\begin{table}
\centering
\begin{tabular}{c|c|c|c}
\hline
& P & R & F1  \\
\hline
RTHN-No-GPE & 0.7369 & 0.7276 & 0.7314 \\
RTHN-No-RPE & 0.4588 & 0.3804 & 0.4145 \\
RTHN-APE & 0.5800 & 0.5618 & 0.5694 \\
RTHN & 0.7697 & 0.7662 & {\bf 0.7677} \\
\hline
\end{tabular}
\caption{The effect of global prediction and different ways of using position information.}
\label{tab:plain}
\end{table}

\begin{table}
\centering
\begin{tabular}{c|c|c|c|c}
\hline
& P & R & F1 & Training Time (s) \\
\hline
RRHN & 0.7831 & 0.7273 & 0.7534 & 732 \\
TTHN & 0.7123 & 0.6798 & 0.6952 & 281\\
RTHN & 0.7697 & 0.7662 & {\bf 0.7677} & 360\\
\hline
\end{tabular}
\caption{Performance of different combinations of RNN and Transformer.}
\label{tab:plain}
\end{table}
\section{Conclusions}
The emotion cause extraction task was normally regarded as a set of independent clause classification problems where the relations between multiple clauses in a document were ignored. In this work, we propose a joint emotion cause extraction framework, called RNN-Transformer Hierarchical Network (RTHN), that can model and classify multiple clauses in a document synchronously. Transformer has demonstrated superior performance in capturing the correlations between multiple clauses. Moreover, we proposed ways to encode two explicit factors in ECE (i.e., relative position and global prediction) that can capture the causality between clauses and make RTHN more efficient for emotion cause extraction. The experimental results on a benchmark ECE corpus verified the effectiveness and superiority of our approach, in comparison with state-of-the-art techniques in ECE.

\section{Acknowledgments}
The work was supported by the Natural Science Foundation of China (No. 61672288), and the Natural Science Foundation of Jiangsu Province for Excellent Young Scholars (No. BK20160085).
\bibliographystyle{named}
\bibliography{ijcai19}

\end{document}